\documentclass[11pt,a4paper]{article}
\usepackage[hyperref]{acl2017}
\usepackage{times}
\usepackage{latexsym}

\usepackage{url}
\usepackage{times}
\usepackage{latexsym}
\usepackage{amssymb,amsmath,amsthm}
\usepackage{multirow}
\usepackage{graphicx}
\usepackage{bm}
\usepackage{makecell}
\usepackage[american]{babel}
\usepackage{balance}
\usepackage{CJK}
\aclfinalcopy 
\def\aclpaperid{2085} 

\title{Fast and Accurate Neural Word Segmentation for Chinese}
\author{Deng Cai$^{1,2}$, Hai Zhao$^{1,2,}$\thanks{\;\;Corresponding author. This paper was partially supported by Cai Yuanpei Program (CSC No. 201304490199 and No. 201304490171), National Natural Science Foundation of China (No. 61170114, No. 61672343 and No. 61272248), National Basic Research Program of China (No. 2013CB329401), Major Basic Research Program of Shanghai Science and Technology Committee (No. 15JC1400103), Art and Science Interdisciplinary Funds of Shanghai Jiao Tong University (No. 14JCRZ04), Key Project of National Society Science Foundation of China (No. 15-ZDA041), and the joint research project with Youtu Lab of Tencent.}\;, Zhisong Zhang$^{1,2}$, Yuan Xin$^3$, Yongjian Wu$^3$, Feiyue Huang$^3$\\ $^1$Department of Computer Science and Engineering, Shanghai Jiao Tong University \\ $^2$Key Lab of Shanghai Education Commission for Intelligent Interaction \\and Cognitive Engineering, Shanghai Jiao Tong Univeristy, Shanghai, China \\ {\tt thisisjcykcd@gmail.com, \{zhaohai@cs, zzs2011@\}sjtu.edu.cn}\\ $^3$Tencent Youtu Lab, Shanghai, China \\{\tt \{macxin,littlekenwu,garyhuang\}@tencent.com}}
\date{}

\begin{document}
	\maketitle
	\begin{abstract}
	Neural models with minimal feature engineering have achieved competitive performance against traditional methods for the task of Chinese word segmentation. However, both training and working procedures of the current neural models are computationally inefficient. This paper presents a greedy neural word segmenter with balanced word and character embedding inputs to alleviate the existing drawbacks. Our segmenter is truly end-to-end, capable of performing segmentation much faster and even more accurate than state-of-the-art neural models on Chinese benchmark datasets. 
	\end{abstract}
	\section{Introduction}
	Word segmentation is a fundamental task for processing most east Asian languages, typically Chinese. Almost all practical Chinese processing applications essentially rely on Chinese word segmentation (CWS), e.g., \cite{zhao2017hybrid}.
	
	Since \cite{xue2003chinese}, most methods formalize this task as a sequence labeling problem. In a supervised learning fashion, sequence labeling may adopt various models such as Maximum Entropy (ME) \cite{low2005maximum} and Conditional Random Fields (CRF) \cite{lafferty2001conditional,peng2004chinese}. However, these models rely heavily on hand-crafted features.
	
	To minimize the efforts in feature engineering, neural word segmentation has been actively studied recently. \newcite{zheng-chen-xu:2013:EMNLP} first adapted the sliding-window based sequence labeling \cite{collobert2011natural} with character embeddings as input. A number of other researchers have attempted to improve the segmenter of \cite{zheng-chen-xu:2013:EMNLP} by augmenting it with additional complexity. \newcite{pei-ge-chang:2014:P14-1} introduced tag embeddings. \newcite{chen-EtAl:2015:ACL-IJCNLP5} proposed to model $n$-gram features via a gated recursive neural network (GRNN). \newcite{chen-EtAl:2015:EMNLP2} used a Long short-term memory network (LSTM) \cite{hochreiter1997long} to capture long-distance context. \newcite{xu-sun:2016:P16-2} integrated both GRNN and LSTM for deeper feature extraction. 
	
	Besides sequence labeling schemes, \newcite{mszhang:2016} proposed a transition-based framework. \newcite{liu2016exploring} used a zero-order semi-CRF based model. However, these two models rely on either traditional discrete features or non-neural-network components for performance enhancement, their performance drops rapidly when solely depending on neural models. Most closely related to this work, \newcite{cai-zhao:2016:P16-1} proposed to score candidate segmented outputs directly, employing a gated combination neural network over characters for word representation generation and an LSTM scoring model for segmentation result evaluation. 
	
	Despite the active progress of most existing works in terms of accuracy, their computational needs have been significantly increased to the extent that training a neural segmenter usually takes days even using cutting-edge hardwares. Meanwhile, different applications often require diverse segmenters and offer large-scale incoming data. The efficiency of a word segmenter either for training and decoding is crucial in practice. In this paper, we propose a simple yet accurate neural word segmenter who searches greedily during both training and working to overcome the existing efficiency obstacle. Our evaluation will be performed on Chinese benchmark datasets.
	\section{Related Work}
	Statistical Chinese word segmentation has been studied for decades \cite{huang2007chinese}. \cite{xue2003chinese} was the first to cast it as a character-based tagging problem.
	\newcite{peng2004chinese} showed CRF based model is particularly effective to solve CWS in the sequence labeling fashion. This method has been followed by most later segmenters \cite{tseng2005conditional,zhao2006effective,zhao2008unsupervised,zhao2010unified,sun2012fast,zhang-EtAl:2013:EMNLP1}. The same spirit has also be followed by most neural models \cite{zheng-chen-xu:2013:EMNLP,pei-ge-chang:2014:P14-1,qi2014deep,chen-EtAl:2015:ACL-IJCNLP5,chen-EtAl:2015:EMNLP2,ma-hinrichs:2015:ACL-IJCNLP,xu-sun:2016:P16-2}.
	
	Word based CWS to conveniently incorporate complete word features has also be explored. \newcite{andrew2006hybrid} proposed a semi-CRF model. \newcite{zhang-clark:2007:ACLMain,zhang2011syntactic} used a perceptron algorithm with inexact search. Both of them have been followed by neural model versions \cite{liu2016exploring} and \cite{mszhang:2016} respectively. There are also works integrating both character-based and word-based segmenters \cite{huang2006essential,sun:2010:POSTERS,wang-voigt-manning:2014:P14-2} and semi-supervised learning \cite{zhao2008exploiting,zhao2011integrating,zeng-EtAl:2013:ACL2013,zhang-EtAl:2013:EMNLP1}.
	
	Unlike most previous works, which extract features within a fixed sized sliding window, \newcite{cai-zhao:2016:P16-1} proposed a direct segmentation framework that extends the feature window to cover complete input and segmentation history and uses beam search for decoding. In this work, we will make a series of significant improvement over the basic framework and especially adopt greedy search instead. 
	
	Another notable exception of embedding based methods is \cite{ma-hinrichs:2015:ACL-IJCNLP}, which used character-specified tags matching for fast decoding and resulted in a character-based greedy segmenter.
	\section{Models}
	To segment a character sequence, we employ neural networks to score the likelihood of a candidate segmented sequence being a true sentence, and the one with the highest score will be picked as output.
	\subsection{Neural Scorer}
\begin{figure}[t]
		\centering
		\scalebox{0.88}[0.88]{
			\includegraphics{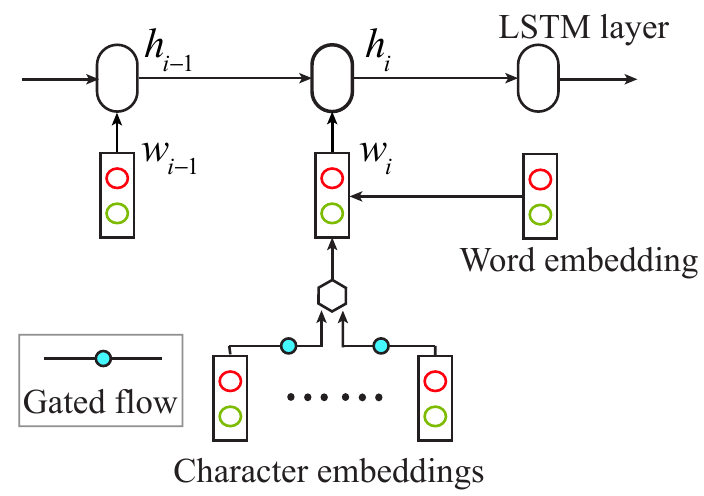}}
		\caption{Neural network scoring for word candidate $w_i$ in a possible word sequence ($...,w_i,...$).}
		\label{arh}
	\end{figure}

	Our neural architecture to score a segmented sequence (word sequence) can be described in the following three steps (illustrated in Figure \ref{arh}).
	
	\paragraph{Encoding}\label{word-character} To make use of neural networks, symbolic data needs to be transformed into distributed representations. The most straightforward solution is to use a lookup table for word vectors \cite{bengio2003neural}. However, in the context of neural word segmentation, it will generalize poorly due to the severe word sparsity in Chinese. An alternative is employing neural networks to compose word representations from character embedding inputs. However, it is empirically hard to learn a satisfactory composition function. In fact, quite a lot of Chinese words, like \begin{CJK*}{UTF8}{gbsn}``沙(sand)发(issue)'' (sofa) \end{CJK*}, are not semantically character-level compositional at all.
	 
	For the dilemma that composing word representations from character may be insufficient while the direct use of word embedding may lose generalization ability, we propose a hybrid mechanism to alleviate the problem. Concretely, we keep a short list $\mathcal{H}$ of the most frequent words $w=c_1..c_l$ to balance character composition. If $w$ in $\mathcal{H}$, the immediate word embedding $\mathbf{w}\in \mathbb{R}^{d_w}$ is attached via average pooling\footnote{We tried other two integration functions, concatenation and adaptive gating mechanism, but it finally shows that the simplest averaging works best.}, otherwise, the character composition is used alone.
	\begin{equation}
		\nonumber
		\textsc{Word}(c_1..c_l) = \begin{cases}
			\frac{\textsc{Comp}(c_1..c_l) + \mathbf{w}[w]}{2} & \text{if } c_1..c_l \in \mathcal{H}\\
			\textsc{Comp}(c_1..c_l) & \text{otherwise}
		\end{cases} 
	\end{equation}
	Our character composition function $\textsc{Comp}(\cdot)$ for $l$-length word is
	\begin{equation}
		\nonumber
		\textsc{Comp}(c_1..c_l)=\tanh(\mathbf{W}^c_l [\mathbf{r}_1\odot\mathbf{c}_1;\ldots;\mathbf{r}_l\odot\mathbf{c}_l] + \mathbf{b}^c_l)
	\end{equation}
	where $\odot$ denotes the element-wise multiplication. $\mathbf{r}_i\in\mathbb{R}^{d_c}$ is the gate that controls the information flow from character embedding $\mathbf{c}_i\in\mathbb{R}^{d_c}$ to word. Intuitively, the gating mechanism is used to determine which part of the character vectors should be retrieved when composing a certain word. This is indeed important due to the ambiguity of individual Chinese characters.
	\begin{equation}
		\nonumber
		[\mathbf{r}_1;\ldots;\mathbf{r}_l] = \text{sigmoid}(\mathbf{W}^r_l [\mathbf{c}_1;\ldots;\mathbf{c}_l] + \mathbf{b}^r_l )
	\end{equation}
	
\begin{figure}[t]
		\centering
		\scalebox{0.88}[0.88]{
			\includegraphics{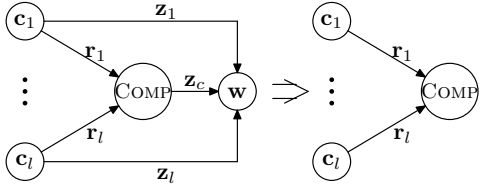}}
		\caption{The difference between \cite{cai-zhao:2016:P16-1} (left) and our model (right).}
		\label{comp}
	\end{figure}

	In contrast, the model in \cite{cai-zhao:2016:P16-1} further combined $\textsc{Comp}(\cdot)$ and character embeddings $\mathbf{c}_i$ via an update gate $\mathbf{z}$ (As in Figure \ref{comp}), which has been shown helpless to the performance but requires huge computational cost according to our empirical study.
	\paragraph{Linking} To capture word interactions within a word sequence, the resulted word vectors are then linked sequentially via an LSTM \cite{sundermeyer2012lstm}. At each time step $i$, a prediction about next word is made according to the current hidden state $\mathbf{h}_i\in\mathbb{R}^H$ of LSTM. The procedure can be described as the following equation.
	\begin{equation}
		\nonumber
		\mathbf{p}_{i+1} = \tanh(\mathbf{W}^p\mathbf{h}_i+\mathbf{b}^p)\\
	\end{equation}
	The predictions $\mathbf{p}\in\mathbb{R}^{d_w}$ will then be used to evaluate how reasonable the transition is between next word and the preceding word sequence.
	
	\paragraph{Scoring} The segmented sequence is evaluated from two perspectives, (i) the legality of individual words, (ii) the smoothness or coherence of the word sequence. The former is judged by a trainable parameter vector $\mathbf{u}\in \mathbb{R}^{d_w}$, which is supposed to work like a hyperplane separating legal and illegal words. For the latter, the prediction $\mathbf{p}$ made for each position can be used to score the fitness of the actual word. Both scoring operations are implemented via dot product in our settings. Summing up all scores, the segmented sequence (sentence) is scored as follow.	
	\begin{equation}
		\nonumber
		s([w_1,w_2,\ldots,w_n]) = \sum_{i=1}^{n} (\mathbf{u} + \mathbf{p}_i)^\text{T}\textsc{Word}(w_i)
	\end{equation}
	\subsection{Search}
	The number of possible segmented sentences grows exponentially with the length of the input character sequence. Most existing methods made Markov assumptions to keep the exact search tractable.\footnote{By assuming that tag interactions or word interactions only exist in adjacent positions.} However, such assumptions cannot be made in our model as the LSTM component takes advantage of the full segmentation history. We then adopt a beam search scheme, which works iteratively on every prefix of the input character sequence, approximating the $k$ highest-scored word sequences of each prefix (i.e., $k$ is the beam size). The time complexity of our beam search is $O(wkn)$, where $w$ is the maximum word length and $n$ is the input sequence length.
	\subsection{Training Criteria}
	Our segmenter is trained using max-margin methods \cite{taskar2005learning} where the structured margin loss is defined as $\mu$ times the number of incorrectly segmented characters \cite{cai-zhao:2016:P16-1}. However,  according to \cite{huang-fayong-guo:2012:NAACL-HLT}, \textbf{standard} parameter update cannot guarantee convergence in the case of inexact search. We thus additionally examine two strategies as follows.
	
	\textbf{Early update} This strategy proposed in \cite{collins-roark:2004:ACL} can be simplified into ``update once the golden answer is unreachable''. In our case, when the considering character prefix can be correctly segmented but the correct one falls off the beam, an update operation will be conducted and the rest part will be ignored.
	
	\textbf{LaSO update} One drawback of early update is that the search may never reach the end of a training instance, which means the rest part of the instance is ``wasted''. Differently, LaSO method of \cite{daume2005learning} continues on the same instance with correct hypothesis after each update. In our case, the beam will be emptied and the corresponding prefix of the correct word sequence will be inserted into the beam.
		\begin{table}[t]
			\centering
			\begin{tabular}{l|r r |r r}
				\hline
				& \multicolumn{2}{c|}{PKU} & \multicolumn{2}{c}{MSR}\\
				\hline
				& Train & Test &Train &Test\\
				\#sentences &19K   & 2K   &87K    &4K \\
				\#words     &1,110K & 104K &2,368K  &107K\\
				\#characters&1,788K & 169K &3,981K  &181K\\  
				\hline
			\end{tabular}
			\caption{Data statistics.}
			\label{data}
		\end{table}
		\begin{table}[t]
			\centering
			\begin{tabular}{l|l}
				\hline
				Character embedding size & $d_c=50$\\
				Word embedding size & $d_w=50$\\
				Hidden unit number & $H=50$ \\ 
				Margin loss discount & $\mu=0.2$\\
				Maximum word length & $w=4$\\
				\hline
			\end{tabular}
			\caption{Model setting.}
			\label{model_setting}
		\end{table}
	\section{Experiments}
	\subsection{Datasets and Settings}
	We conduct experiments on two popular benchmark datasets, namely PKU and MSR, from the second international Chinese word segmentation bakeoff \cite{emerson2005second} (Bakeoff-2005). Data statistics are in Table \ref{data}.
	
	Throughout this paper, we use the same model setting as shown in Table \ref{model_setting}. These numbers are tuned on development sets.\footnote{Following conventions, the last 10\% sentences of training corpus are used as development set.} We follow \cite{dyer-EtAl:2015:ACL-IJCNLP} to train model parameters. The learning rate at epoch $t$ is set as $\eta_t = 0.2/(1+\gamma t)$, where $\gamma=0.1$ for PKU dataset and $\gamma=0.2$ for MSR dataset. The character embeddings are either randomly initialized or pre-trained by word2vec \cite{mikolov2013efficient} toolkit on Chinese Wikipedia corpus (which will be indicated by \textit{+pre-train} in tables.), while the word embeddings are always randomly initialized. The beam size is kept the same during training and working. By default, early update strategy is adopted and the word table $H$ is top half of in-vocabulary (IV) words by frequency.
	\subsection{Model Analysis}
			\begin{figure}[t]
				\centering
				\scalebox{1.}[1.]{
					\includegraphics{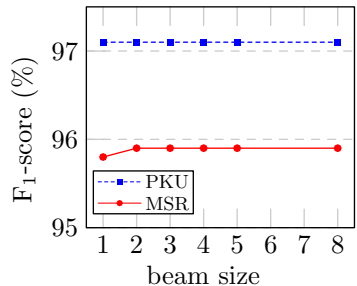}}
				\caption{The effect of different beam sizes.}
				\label{beam}
			\end{figure}
			\begin{table}[t]
				\centering			
				\begin{tabular}{c|cc|cc}
					\hline
					\multirow{2}{*}{Methods}& \multicolumn{2}{c|}{PKU} & \multicolumn{2}{c}{MSR}\\
					& $\text{F}_1$  & \#epochs & $\text{F}_1$ & \#epochs\\
					\hline
					Standard & 95.6 & 50 &96.7 & 50\\
					Early update & \textbf{95.8}&\textbf{30}& \textbf{97.1}&\textbf{30}\\
					LaSO update &95.7 &45& 97.0&30\\
					\hline
				\end{tabular}
				\caption{The effect of different update methods. \#epochs denotes the number of training epochs to convergence.}
				\label{update_method}
			\end{table}
	\paragraph{Beam search collapses into greedy search}
	Figure \ref{beam} demonstrates the effect of beam size. To our surprise, beam size change has little influence on the performance. Namely, simple step-wise greedy search nearly achieves the best performance, which suggests that word segmentation can be greedily solvable at word-level. It may be due to that right now the model is optimal enough to make correct decisions at the first position. In fact, similar phenomenon was observed at character-level \cite{ma-hinrichs:2015:ACL-IJCNLP}. The rest experiments will thus only report the results of our greedy segmenter.
	\begin{figure}[t]
		\centering
		\scalebox{1.}[1.]{
			\includegraphics{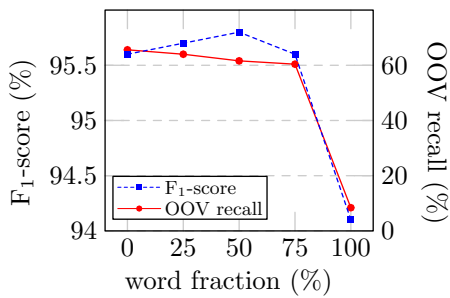}}
		\caption{Performance with different sizes of word table on PKU test set.}
		\label{pku}
	\end{figure}
		\begin{table*}[t]
			\small
			\centering
			\begin{tabular}{c|cccc|cccc}
				\hline
				\multirow{3}{*}{Models}& \multicolumn{4}{c|}{PKU} &\multicolumn{4}{c}{MSR}\\
				\cline{2-9}
				& \makecell[c]{$\text{F}_1$\\ + \textit{pre-train}} & \makecell[c]{$\text{F}_1$\\ } &\makecell[c]{Training\\(hours)}&\makecell[c]{Test \\(sec.) }&\makecell[c]{$\text{F}_1$\\ + \textit{pre-train}} & \makecell[c]{$\text{F}_1$\\ } &\makecell[c]{Training\\(hours)}&\makecell[c]{Test \\(sec.) }\\
				\hline
				\cite{zhao2008unsupervised} & -  & 95.4 & - & - & - & \textbf{97.6}  & - & -\\
				\hline
				\cite{chen-EtAl:2015:ACL-IJCNLP5}&  94.5* &94.4* & 50 &105 & 95.4*  & 95.1* & 100&120\\
				\cite{chen-EtAl:2015:EMNLP2}& 94.8* &94.3* &  58 & 105 &  95.6*  &95.0* &117&120\\
				\cite{ma-hinrichs:2015:ACL-IJCNLP}& -&95.1& \textbf{1.5} & \textbf{24} &- &96.6& \textbf{3} & \textbf{28} \\
				\cite{mszhang:2016}& 95.1 & - & 6 & 110 & 97.0 & -  & 13 & 125\\
				\cite{liu2016exploring}& 93.91  & - & - & - & 95.21 & -  & - & -\\
				\cite{cai-zhao:2016:P16-1}& 95.5 & 95.2 & 48 & 95 & 96.5  & 96.4  & 96&105\\
				\hline
				Our results & \textbf{95.8} & \textbf{95.4} & 3&25  & \textbf{97.1} & 97.0 &6&30\\
				\hline
			\end{tabular}
			\caption{Comparison with previous models. Results with * are from \cite{cai-zhao:2016:P16-1}.\protect\footnotemark[4]}
			\label{final result}
		\end{table*}

	\paragraph{Comparing different update methods} Table \ref{update_method} compares the concerned three training strategies. We find that early update leads to faster convergence and a bit better performance compared to both standard and LaSO update.
	
	\paragraph{Character composition versus word embedding}
	Following Section \ref{word-character}, direct use of word embedding may bring efficiency and effectiveness for identifying IV words, but weaken the ability to recognize out-of-vocabulary (OOV) words. We accordingly conduct experiments on different sizes of word table $\mathcal{H}$. Concretely, sorting all IV words by frequency, the first \{0, 25\%, 50\%, 75\%, 100\%\} fraction of them respectively forms the word table. The corresponding results on PKU in Figure \ref{pku} demonstrate that by the use of direct word embedding, $\text{F}_1$ score increases first but then drops rapidly. In contrast, OOV recall, which partially reflects the model generalization ability, decreases consistently. In addition, we also found the number of training epochs to convergence decreases continually. Overall, the results indicate that word-aware segmenter learns faster and fits better on training set, but generalizes relatively poorly.
	\subsection{Main Results}
	Table \ref{final result} compares our final results (greedy search is adopted by setting $k$=1) to prior neural models. Pre-training character embeddings on large-scale unlabeled corpus (not limited to the training corpus) has been shown helpful for extra performance improvement. The results with or without pre-trained character embeddings are listed separately for following the strict closed test setting of SIGHAN Bakeoff in which no linguistic resource other than training corpus is allowed. We also show the state-of-the-art results in (Zhao and Kit, 2008b) of traditional methods.\footnotetext[4]{To distinguish the performance improvement from model optimization, we especially list the results of stand-alone neural models in \cite{mszhang:2016} and \cite{liu2016exploring}. All the running time results are from our runs of released implementations on a single CPU (Intel i7-5960X) with two threads only, except for those of \cite{mszhang:2016} which are from personal communication. The results of \cite{xu-sun:2016:P16-2} are not listed due to their use of external Chinese idiom dictionary.} The comparison shows our neural word segmenter outperforms all state-of-the-art neural systems with much less computational cost. 
	
	Finally, we present the results on all four Bakeoff-2005 datasets compared to \cite{zhao2008unsupervised} in Table \ref{four}. Note \cite{zhao2008unsupervised} used AV features, which are derived from global statistics over entire training corpus in a similar way of unsupervised segmentation \cite{zhao2008empirical}, for performance enhancement.\footnote[5]{In fact, this kind of features may also be incorporated to our model. We leave it as future work.} The comparison to their results without AV features show that our neural models for the first time present comparable performance against state-of-the-art traditional ones under strict closed test setting.\footnote[6]{To our knowledge, none of previous neural models has ever performed a complete evaluation over all four segmentation corpora of Bakeoff-2005, in which only two, PKU and MSR, are used since \cite{pei-ge-chang:2014:P14-1}.}
	\begin{table}[t]
		\centering
		\small
		\begin{tabular}{c|cccc}
			\hline
			Models&PKU & MSR & CityU & AS\\
			\hline
			\cite{zhao2008unsupervised}&95.4  & 97.6 & 96.1 & 95.7\\
			--\textit{AV} &95.2 &\textbf{97.4}& 94.8 &\textbf{95.3}\\
			\hline
			ours            &\textbf{95.4}  & 97.0 & \textbf{95.4} & 95.2\\
			+\textit{pre-train}&95.8  & 97.1 & 95.6 & 95.3\\
			\hline
		\end{tabular}
		\caption{Results on all four Bakeoff-2005 datasets.}
		\label{four}
	\end{table}
	\section{Conclusion}
	In this paper, we presented a fast and accurate word segmenter using neural networks. Our experiments show a significant improvement over existing state-of-the-art neural models by adopting the following key model refinements.

	(1) A novel character-word balanced mechanism for word representation generation. (2) A more efficient model for character composition by dropping unnecessary designs. (3) Early update strategy during max-margin training. (4) With the above modifications, we discover that beam size has little influence on the performance. Actually, greedy search achieves very high accuracy.
	
	Through these improvement from both neural models and linguistic motivation, our model becomes simpler, faster and more accurate.\footnote[7]{Our code based on Dynet \cite{dynet} is released at \url{https://github.com/jcyk/greedyCWS}.}
	\balance
	\bibliography{acl2017}
	\bibliographystyle{acl_natbib}
\end{document}